\documentclass[runningheads]{llncs}

\usepackage{eccv}

\usepackage{eccvabbrv}

\usepackage{graphicx}
\usepackage{booktabs}
\usepackage{wrapfig}
\usepackage{multirow}
\usepackage{colortbl}
\usepackage{xcolor}
\usepackage{enumitem}
\usepackage{tcolorbox}
\usepackage{fontawesome5}

\usepackage[accsupp]{axessibility} 

\usepackage[pagebackref, breaklinks, colorlinks, citecolor=eccvblue]{hyperref}

\usepackage{orcidlink}

\usepackage{pifont}
\newcommand{\cmark}{\ding{51}}
\newcommand{\xmark}{\ding{55}}
\newcommand{\pmark}{\ding{218}}

\definecolor{MainPurple}{RGB}{102, 0, 153}

\begin{document}
    \title{AmodalSVG: Amodal Image Vectorization via Semantic Layer Peeling}


    \author{
        Juncheng Hu\inst{1} \and 
        Ziteng Xue\inst{1} \and 
        Guotao Liang\inst{1} \and
        Anran Qi\inst{2} \and \\
        Buyu Li\inst{3} \and
        Sheng Wang\inst{3} \and
        Dong Xu\inst{4} \and
        Qian Yu\inst{1}\textsuperscript{\faEnvelope[regular]}
    }
    {
      \renewcommand{\thefootnote}{\faEnvelope[regular]}
      \footnotetext{ Corresponding author.}
    }
    \authorrunning{J.~Hu et al.}

    \institute{
        School of Software, Beihang University \\
        \email{\{hujuncheng, zt\_xue, liangguotao, qianyu\}@buaa.edu.cn} \and
        Igarashi Lab, The University of Tokyo \\
        \email{annranqi@g.ecc.u-tokyo.ac.jp} \and
        Bambu Lab \\
        \email{libuyu\_braiten@outlook.com, sheng.wang@pku.edu.cn} \and
        Department of Computer Science, The University of Hong Kong \\
        \email{dongxu@cs.hku.hk}
    }

    \maketitle

    \begin{figure}
        \centering
        \includegraphics[width=\linewidth]{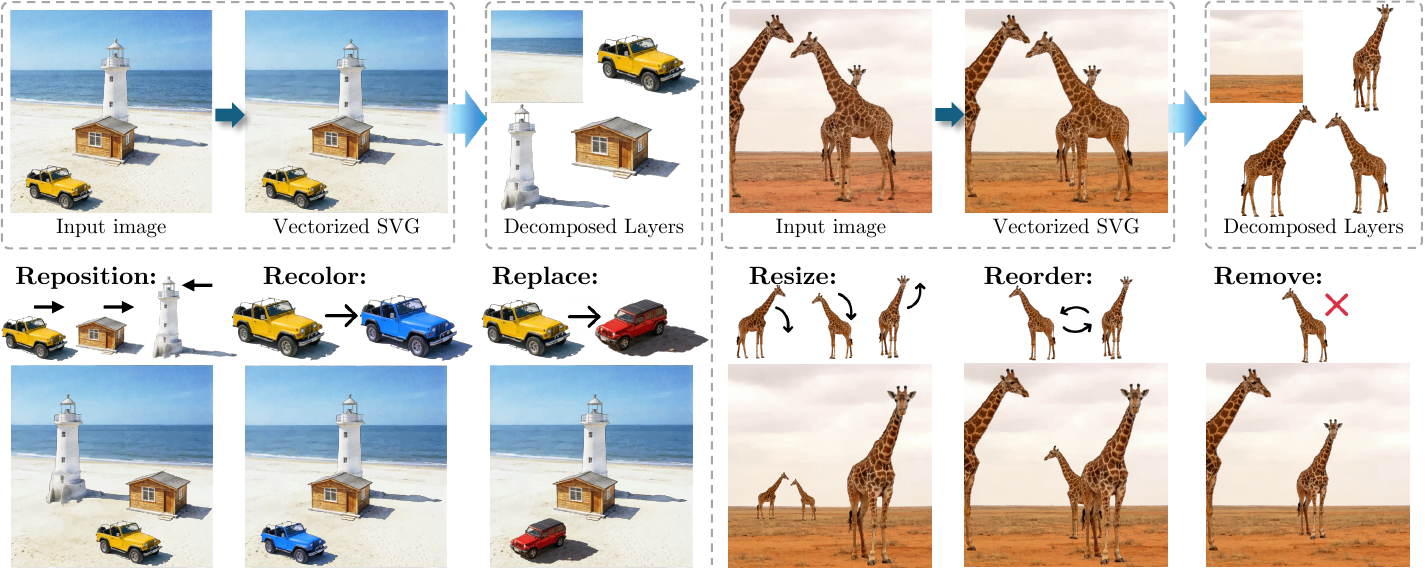}
        \vspace{-2em}
        \caption{\textbf{AmodalSVG} vectorizes a raster image into semantically decoupled and amodally complete vector layers (\eg, the unoccluded lighthouse and its shadow on the top left, and the full giraffe on the top right). The resulting vector layers enable object-level editing--such as \textit{Reposition}, \textit{Recolor}, \textit{Replace}, \textit{Resize}, \textit{Reorder}, and \textit{Remove}--directly in the vector domain while maintaining global visual consistency (bottom).}
        \vspace{-3em}
        \label{fig:teaser}
    \end{figure}

    \begin{abstract}
    We introduce \textbf{AmodalSVG}, a new framework for \textbf{amodal image vectorization} that produces semantically organized and geometrically complete SVG representations from natural images. Existing vectorization methods operate under a modal paradigm: tracing only visible pixels and disregarding occlusion. Consequently, the resulting SVGs are semantically entangled and geometrically incomplete, limiting SVG's structural editability. In contrast, AmodalSVG reconstructs full object geometries, including \textit{occluded} regions, into independent, editable vector layers. To achieve this, AmodalSVG reformulates image vectorization as a two-stage framework, performing semantic decoupling and completion in the raster domain to produce amodally complete semantic layers, which are then independently vectorized. In the first stage, we introduce \textbf{Semantic Layer Peeling (SLP)}, a VLM-guided strategy that progressively decomposes an image into semantically coherent layers. By hybrid inpainting, SLP recovers complete object appearances under occlusions, enabling explicit semantic decoupling. To vectorize these layers efficiently, we propose \textbf{Adaptive Layered Vectorization (ALV)}, which dynamically modulates the primitive budget via an error-budget-driven adjustment mechanism. Extensive experiments demonstrate that AmodalSVG significantly outperforms prior methods in visual fidelity. Moreover, the resulting amodal layers enable object-level editing directly in the vector domain—capabilities not supported by existing vectorization approaches. Code will be released upon acceptance.



\end{abstract}

    \section{Introduction}
Vector graphics (\eg, SVG)~\cite{svg_w3c_1999} are foundational to digital art, graphic design, and animation due to their resolution independence and structural editability. However, since the vast majority of digital visual content is created and stored as raster images, there is a strong practical demand to convert these pixel grids into vector formats. Consequently, image vectorization—the task of translating raster images into parametric geometric primitives—has remained an active research topic for decades, witnessing substantial progress in recent years.



Despite achieving impressive reconstruction fidelity, existing frameworks~\cite{potrace_selinger_2003, vtracer_visioncortex_2020, diffvg_li_2020, im2vec_reddy_2021, optimizeandreduce_hirschorn_2024} predominantly operate under a \textit{modal} perception paradigm: they trace only the visible pixels and ignore the occluded regions. In real-world scenes containing occlusions, these methods fail to achieve meaningful hierarchical organization. When an object is partially occluded, modal vectorization splits it into disconnected, meaningless vector fragments rather than a cohesive entity. While path-by-path layering methods~\cite{adavec_zhao_2025, layerpeeler_wu_2025} can successfully recover layers during vectorization, they are specifically designed for flat icons or emojis and are difficult to extend to natural images. Some recent methods attempt coarse-to-fine layering~\cite{live_ma_2022, sglive_zhou_2024, layertracer_song_2025, layered_wang_2025}; however, these layers rarely correspond to distinct semantic entities. Consequently, the resultant vector graphics from existing methods cannot support object-level isolation, repositioning, or manipulation, which significantly restricts their downstream applications.

To address this limitation, we introduce \textbf{\textit{Amodal} Image Vectorization}, a novel task that aims at recovering the complete geometry and appearance of objects, including their occluded regions. Due to the lack of strong generative priors necessary to infer missing geometry and disentangle intertwined primitives, completing an object in the vector domain is exceedingly difficult. Conversely, the raster domain benefits from powerful Vision-Language Models (VLMs) for scene understanding and generative diffusion models for image completion.

Motivated by this insight, we propose \textbf{AmodalSVG}, a framework designed to achieve amodal vectorization by performing semantic decoupling and completion in the raster space prior to vectorization. Specifically, we introduce \textbf{Semantic Layer Peeling} (\textbf{SLP}), which iteratively identifies and ``peels'' off semantic layers from foreground to background guided by VLMs. By treating occlusions as missing data and applying hybrid inpainting in the raster domain, SLP recovers the full, amodal appearance of each object. Vectorization is performed only after these complete, independent raster layers are constructed. This pipeline ensures that the final vector graphics consists of semantically separated, amodally complete layers, enabling flexible downstream editing such as individual recoloring, resizing, repositioning, and removal.

Furthermore, managing primitive complexity when vectorizing individual semantic layers remains challenging. We propose \textbf{Adaptive Layered Vectorization} (\textbf{ALV}), which treats primitive management as a dynamic, feedback-driven process. Unlike prior methods that rely on fixed primitive schedules~\cite{diffvg_li_2020, live_ma_2022, im2vec_reddy_2021} or heuristic initialization~\cite{vtracer_visioncortex_2020, samvg_zhu_2024, optimizeandreduce_hirschorn_2024}, ALV adaptively allocates primitives through a feedback-driven strategy that combines occlusion-aware primitive pruning with importance-driven primitive addition. 
This mechanism dynamically controls the number of primitives, ensuring that every layer is represented by a compact set of primitives that captures intricate details while preventing over-parameterization. Consequently, it achieves the balance between reconstruction fidelity and primitive efficiency.

Our main contributions are summarized as follows:
\begin{itemize}
    \item We introduce the task of Amodal Image Vectorization. While preserving the fundamental objective of raster-to-vector conversion, this new task necessitates the holistic reconstruction of complete object geometries, including occluded regions. We accordingly propose AmodalSVG, a novel framework that strategically performs raster-level semantic decoupling and completion prior to layer-wise vectorization.


    \item For semantic decoupling and completion, we introduce Semantic Layer Peeling (SLP), a VLM-guided iterative strategy that progressively extracts semantic layers and utilizes hybrid inpainting to recover the complete, amodal appearances of occluded objects.

    \item For layer-wise vectorization, we propose Adaptive Layered Vectorization (ALV), a feedback-driven mechanism that dynamically controls the number of primitives based on regional complexity, ensuring each semantic layer is represented by an optimal and concise set of primitives.


    \item Extensive experiments demonstrate that AmodalSVG achieves superior visual fidelity over state-of-the-art image vectorization methods. Moreover, it uniquely produces semantically organized, amodally complete vector layers that enable broader object-level editings in the vector domain.
\end{itemize}

    \section{Related works}

\subsection{Vector Graphics Generation}
Vector graphics generation has evolved from unconditional synthesis~\cite{sketchrnn_david_2018, deepsvg_carlier_2020} to conditional paradigms, driven by image~\cite{clipasso_vinker_2022, wordasimage_iluz_2023, clipascene_vinker_2023} or text~\cite{clipdraw_frans_2022, wordasimage_iluz_2023, vectorfusion_jain_2023, diffsketcher_xing_2023, svgdreamer_xing_2024, svgdreamer++_xing_2025}. Recently, (Multimodal) Large Language Models (LLMs/MLLMs)~\cite{sketchagent_vinker_2025, llm4svg_xing_2025, chat2svg_wu_2025, svgen_wang_2025, svgthinker_chen_2025, omnisvg_yang_2025, starvector_rodriguez_2025, reasonsvg_xing_2025, rlrf_rodriguez_2025, internsvg_wang_2025, duetsvg_zhang_2025} have been widely employed to directly predict SVG code. However, these generative approaches are largely restricted to simple, flat icons and fail to generalize well to real-world natural imagery.

\subsection{Image Vectorization}
Image vectorization has transitioned from traditional graphics tracing~\cite{potrace_selinger_2003, diffusion_curves_orzan_2008, vtracer_visioncortex_2020} and learning-based regression~\cite{deepsvg_carlier_2020, im2vec_reddy_2021} to formulations based on differentiable rendering and \textit{global optimization}~\cite{diffvg_li_2020,optimizeandreduce_hirschorn_2024}. To introduce structural organization, some work adopts \textit{coarse-to-fine layering} strategies~\cite{live_ma_2022,sglive_zhou_2024,layertracer_song_2025,layered_wang_2025}, refining vector primitives through hierarchical optimization. However, the resulting layers largely function as optimization artifacts rather than semantically decoupled scene entities. Concurrently, methods~\cite{supersvg_hu_2024, samvg_zhu_2024} incorporating spatial grouping priors (\eg, superpixels~\cite{slic_achanta_2012} or SAM~\cite{sam_kirillov_2023}) frequently suffer from over-segmentation in complex scenes.
More recent approaches explore \textit{path-by-path layering}~\cite{layerpeeler_wu_2025, adavec_zhao_2025}, which sequentially extracts vector primitives. While effective for simple icons, they often struggle with complex images, producing fragmented or incomplete layers, as they fail to account for the intricate amodal completion required to recover entire occluded objects. In contrast, AmodalSVG reconstructs complete object geometries, including occluded regions into independent semantic layers, followed by layer-wise vectorization, enabling structurally coherent vector representations for object-level editing.



\subsection{Amodal Perception}
Human vision and recognition systems naturally have the capacity for mentally reconstructing the complete semantic concept of occluded objects, a phenomenon known as amodal perception~\cite{kellman2013perceptual}. Inspired by this capability, significant efforts have been made to enable machines to reason about occluded content in raster-based tasks, such as instance segmentation~\cite{amodal_qi_2019, 2damodal_li_2022, coarse_gao_2023, pix2gestalt_ozguroglu_2024}, appearance completion~\cite{amodal_xu_2024, open_ao_2025}, depth estimation~\cite{amodaldepthanything_li_2025} and autonomous driving~\cite{amodal_qi_2019, corrbev_xue_2025}. We argue that this need stems from the flat nature of pixel-based representations, which inherently lack the structural hierarchy required to represent overlapping pixels. In contrast, scalable vector graphics can naturally encode layered composition, where each object can maintain its own complete semantic integrity in the rendering stack. However, recent vectorization methods overlook the visual content behind occlusion and mainly focus on the visible parts. This motivates us to explore an occlusion-aware framework that explicitly models amodal perception in the context of image vectorization.

    \section{Methodology}

\begin{figure}[t]
    \centering
    \includegraphics[width=\linewidth]{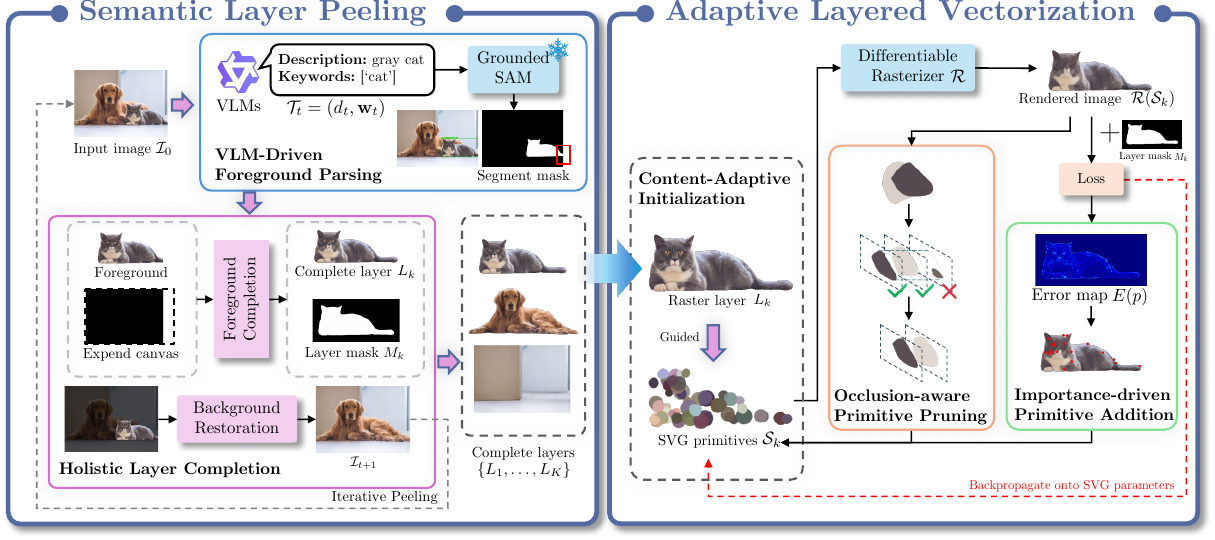}
    \vspace{-2em}
    \caption{\textbf{Overview of the AmodalSVG.} Our framework adopts a two-stage pipeline: raster-level semantic decoupling and completion, followed by layer-wise vectorization. (Left) For semantic decoupling and completion, \textbf{Semantic Layer Peeling (SLP)} iteratively identifies entities via VLMs to guide SAM segmentation. A hybrid inpainting mechanism then reconstitutes occluded regions to produce structurally integral raster layers $L_{t}$. (Right) For layer-wise vectorization, \textbf{Adaptive Layered Vectorization (ALV)} vectorizes each layer $L_{k}$ starting from a content-adaptive initialization. It dynamically manages SVG primitives $\mathcal{S}_{k}$ by pruning redundant primitives via visibility analysis and inserting new primitives based on the error map $E(p)$.}
    \vspace{-1.5em}
    \label{fig:pipeline}
\end{figure}

Given an input raster image $\mathcal{I}$, our goal is to perform \textbf{Amodal Image Vectorization} to produce an SVG representation $\mathcal{S}= \{\mathcal{S}_{1}, \mathcal{S}_{2}, \ldots, \mathcal{S}_{K}\}$. Each layer $\mathcal{S}_{k}$ must correspond to a semantically coherent and amodally complete object, encompassing both its visible and occluded regions. Our \textbf{AmodalSVG} framework achieves this through a synergistic two-stage pipeline: raster-level semantic decoupling and completion, followed by layer-wise vectorization. Specifically, in the first stage, we introduce \textbf{Semantic Layer Peeling (SLP)} (\cref{fig:pipeline} left, \cref{subsec:slp}). SLP iteratively extracts semantic entities and restores occluded regions to produce amodally complete raster layers. In the subsequent layer-wise vectorization stage, we present \textbf{Adaptive Layered Vectorization (ALV)} (\cref{fig:pipeline} right, \cref{subsec:alv}), which vectorizes each layer with a feedback-driven primitive allocation mechanism.

\subsection{Semantic Layer Peeling}
\label{subsec:slp} The core of our method is an iterative peeling process that decomposes the input image $\mathcal{I}$ into a sequence of ordered semantic layers. We formulate this scene decomposition as a recursive problem: at each iteration $t$, the current residual image $\mathcal{I}_{t}$ (where $\mathcal{I}_{0}= \mathcal{I}$) is analyzed to disentangle the foremost semantic entity $L_{t}$ from the background. This entity is then extracted, and the occluded region is inpainted with content that matches the underlying background to form $\mathcal{I}_{t+1}$. This cycle repeats until the scene is reduced to a pure background layer. By explicitly modeling occlusion relationships and restoring hidden contexts, SLP ensures that each extracted layer is not only a collection of surface pixels, but a structurally complete object ready for vectorization.

\subsubsection{(1) VLM-Driven Foreground Parsing.}
To effectively identify the ``foreground'', the system must infer the \textbf{occlusion hierarchy} from 2D pixel arrays. We leverage the emergent spatial reasoning capabilities of advanced Vision-Language Models (VLMs) to perform this depth-aware analysis.

\paragraph{Semantic Depth Reasoning.}
Rather than relying on standard object detectors (\eg, DETR~\cite{detr_carion_2020} or YOLO~\cite{yolo_redmon_2016}) that treat all instances equally, we utilize a VLM to explicitly reason about occlusion orders. Specifically, we design a chain-of-thought prompt for Qwen3-VL~\cite{qwen3vl_bai_2025}, directing the model to analyze spatial overlaps and identify the object closest to the viewer. The VLM outputs a \textit{semantic tuple} $\mathcal{T}_{t}= (d_{t}, \mathbf{w}_{t})$, comprising a dense visual description $d_{t}$ and a set of discriminative keywords $\mathbf{w}_{t}=\{w_{1}, w_{2}, \ldots\}$. Furthermore, the VLM serves as a dynamic termination controller; when it detects no distinct objects superimposed on the background context (\eg, a pure landscape or wall), it issues a termination signal, naturally concluding the peeling process.

\paragraph{Grounded Segmentation.}
Translating the semantic tuple $\mathcal{T}_{t}$ into a pixel-accurate binary mask $M_{t}$ requires bridging the gap between high-level semantics and low-level visual features. To avoid the segmentation fragmentation often caused by unconditional models~\cite{sam_kirillov_2023}, we employ a cascaded grounding approach utilizing GroundedSAM~\cite{groundedsam_ren_2024}. Specifically, the keywords $\mathbf{w}_{t}$ serve as linguistic anchors to generate precise bounding boxes, which subsequently prompt a GroundedSAM yielding a semantically coherent and spatially accurate mask $M_{t}$.

\subsubsection{(2) Holistic Layer Completion.}
To ensure that every extracted layer represents a structurally complete entity, we introduce a holistic completion process that addresses both the foreground object $L_{t}$ and the revealed background $\mathcal{I}_{t+1}$.

\paragraph{Foreground Completion.}
To ensure the extracted foreground object $L_{t}$ is structurally complete even when truncated by the image frame, we introduce a dynamic canvas padding mechanism. Specifically, if the mask $M_{t}$ touches the image boundary, the canvas is expanded outward, providing sufficient spatial room to synthesize the missing parts of the foreground entity via inpainting.

\paragraph{Background Restoration.}
Extracting the foreground object layer $L_{t}$ leaves a missing region (hole) defined by $M_{t}$. While some attempts directly vectorize visible disjoint patches via visible-region segmentation, circumventing the occluded area, such region-wise operations intrinsically cause ``cut-out'' artifacts~\cite{supersvg_hu_2024, samvg_zhu_2024, easy_chen_2025}. Consequently, underlying objects extracted in subsequent iterations become fragmented and lose semantic integrity. To simulate a true peeling process, each layer must remain a structurally complete entity. We achieve this by actively inpainting the missing region with content that matches the underlying background structure. Specifically, we employ hybrid inpainting pipeline that combines LaMa~\cite{lama_suvorov_2022} and FLUX~\cite{flux_fill_blackforestlab_2024} to synthesize this missing information and generate a plausible $\mathcal{I}_{t+1}$. This synergistic approach yields superior results compared to using either model individually, as LaMa excels at efficiently reconstructing background textures and background materials (\eg, brick walls or grass), while FLUX is better suited for semantically rich textures and high-fidelity details (\eg, completing an occluded piece of an object). See the supplementary material for more details.

\subsection{Adaptive Layered Vectorization}
\label{subsec:alv} Given the independent raster layers $\{L_{1}, \ldots, L_{K}\}$ and their restored spatial masks $\{M_{1}, \ldots, M_{K}\}$ extracted via SLP, we aim to individually vectorize each $L_{k}$ into a compact set of SVG primitives $\mathcal{S}_{k}$. To optimally balance reconstruction fidelity and primitive efficiency, we introduce the \textbf{Adaptive Layered Vectorization (ALV)} strategy. Rather than relying on fixed heuristics, ALV dynamically modulates the primitive budget starting from a \textit{content-adaptive initialization}, and iteratively refines the SVG structure through a feedback-driven mechanism of \textit{occlusion-aware pruning} and \textit{importance-driven addition}.

\subsubsection{(1) Content-Adaptive Initialization.}
Since optimizing SVG parameters is a highly non-convex problem~\cite{diffvg_li_2020}, providing a good starting point is crucial. In our approach, we initialize the SVG primitives based on the spatial and color properties of the target layer $L_{k}$. Specifically, we set the initial primitive budget proportional to the area of the mask $M_{k}$, ensuring appropriate representation capacity for each layer. Furthermore, following O\&R~\cite{optimizeandreduce_hirschorn_2024}, the initial positions and colors of primitives are directly sampled from corresponding pixels in the raster image. This targeted initialization prevents primitives from scattering randomly, significantly reducing the parameter's search space.

\subsubsection{(2) Occlusion-aware Primitive Pruning.}
Conventional pruning heuristics often rely on isolated attributes like area or opacity~\cite{svgdreamer++_xing_2025}, which frequently fail in complex scenes. For instance, \textit{area-based pruning} tends to eliminate small-scale primitives which could be crucial for fine-grained details, while \textit{opacity-based pruning} disrupts sophisticated effects (\eg, soft shadows) formed by stacking semi-transparent primitives (see the supplementary material for more analysis).

We notice that prior methods ignore the sequential rendering nature of SVGs, where a primitive's actual \textit{visual contribution} is inextricably linked to its occlusion by overlying primitives. Inspired by redundancy reduction techniques in 3D Gaussian Splatting~\cite{reducing_papantonakis_2024}, we propose using this \textit{visual contribution} within the global occlusion hierarchy for pruning. This ensures that fully occluded primitives are correctly identified as redundant regardless of their opacity, while observable small or semi-transparent primitives are preserved to maintain high visual fidelity.

\paragraph{Theoretical Optimum:}
The ideal measure of visual contribution for the $i$-th primitive $s_{i}$ corresponds to its marginal contribution to the global reconstruction error, \ie, $\Delta \mathcal{L}_{i}= \| \mathcal{R}(\mathcal{S}_{k}\setminus \{s_{i}\}) - L_{k}\| - \| \mathcal{R}(\mathcal{S}_{k}) - L_{k}\|$ (or approximated as $\Delta \mathcal{L}_{i}' \approx \| \mathcal{R}(\mathcal{S}_{k}) - \mathcal{R}(\mathcal{S}_{k}\setminus \{s_{i}\}) \|$), where $\mathcal{R}$ is the rendering function and $\mathcal{S}_{k}$ is the set of primitives. Although theoretically optimal, computing this metric for all primitives requires $N$ separate rendering passes. This leads to a computationally prohibitive $\mathcal{O}(N^{2})$ complexity, making it intractable for complex scenes with massive primitives.

\paragraph{Efficient Proxy:}
To bypass this computational bottleneck, we derive an efficient occlusion-aware proxy based on the rendering hierarchy. Let $\Omega$ denote the discrete image domain (\ie, the set of all pixel coordinates). For each primitive $s_{i}$, we define $\alpha_{i}(p) \in [0, 1]$ as its alpha mask value at pixel $p \in \Omega$, where $0$ indicates full transparency and $1$ indicates full opacity. We recursively compute the \textbf{Cumulative Occlusion} $O_{i}(p) \in [0, 1]^{H \times W}$, which represents the aggregate opacity of all \textit{overlying} primitives $\{i+1, \dots, N\}$ obscuring primitive $s_{i}$:
\vspace{-2pt}
\begin{equation}
    O_{i-1}(p) = O_{i}(p) + \alpha_{i}(p) \cdot (1-O_{i}(p))
\end{equation}
where $O_{N}(p) = \mathbf{0}$, as the topmost primitive remains unoccluded. The effective \textit{visual contribution score} $\mathcal{C}_{i}$ is then formulated as the total visible alpha area of the primitive in the final composition:
\vspace{-4pt}
\begin{equation}
    \mathcal{C}_{i}= \sum_{p \in \Omega}\alpha_{i}(p) \cdot (1 - O_{i}(p))
\end{equation}
Conceptually, this metric quantifies the portion of primitive $s_{i}$ that survives alpha compositing, directly reflecting its absolute pixel-level contribution to the final rendered image. By traversing layers top-to-bottom, all contribution scores $\{\mathcal{C}_{i}\}$ can be computed in a single $\mathcal{O}(N)$ pass, requiring each primitive to be rendered only once. This achieves a dramatic speedup over the differential approach while strictly adhering to the structural logic of vector rendering. Specifically, we prune primitives whose contribution scores $\mathcal{C}_{i}$ fall below a predefined threshold $\tau_{p}$, ensuring only primitives with significant visual impact are retained.

\subsubsection{(3) Importance-driven Primitive Addition.}
To refine regions of insufficient fidelity, we introduce an adaptive primitive addition strategy so that new primitives are concentrated around these areas. Instead of random placement, we leverage the \textbf{residual error field} $E(p)$ (derived from the \texttt{grad-l2} map) to guide the spatial distribution of new primitives. We compute a sampling probability $\mathcal{P}(p)$ via temperature-scaled softmax:
\vspace{-4pt}
\begin{equation}
    \mathcal{P}(p) = \dfrac{E(p)^{1/T}}{\sum_{q \in \Omega} E(q)^{1/T}},
\end{equation}
where $T$ controls the trade-off between exploiting high-error regions (\eg, edges) and exploring broader areas. New primitive coordinates are obtained by multinomial sampling from $\mathcal{P}(p)$. 

\paragraph{Adaptive Quantity via Error Budget.}
A critical challenge in adaptive vectorization is determining \textit{how many} primitives to add. Fixed-increment strategies often lead to either slow convergence or redundant parameterization. We address this by introducing a feedback-driven \textbf{Error Budget} estimation that treats each primitive as a ``resource'' with an associated ``cost-effectiveness''. We maintain a sliding window of the $N$ most recent addition events to calculate the \textit{historical marginal loss reduction} $\Delta e_{add}$:
\begin{equation}
    \Delta e_{add}= \dfrac{1}{N}\sum_{i=1}^{N}\dfrac{\mathcal{L}_{before}^{(i)} - \mathcal{L}_{after}^{(i)}}{n_{added}^{(i)}}
\end{equation}
where $\Delta e_{add}$ represents the empirical average contribution of a single primitive toward minimizing the loss $\mathcal{L}$ (detailed subsequently in Eq.~\ref{eq:total_loss}). Here, $\mathcal{L}_{before}^{(i)}$ and $\mathcal{L}_{after}^{(i)}$ denote the loss values before and after the $i$-th addition of $n_{added}^{(i)}$ primitives. Given the current reconstruction loss $\mathcal{L}_{curr}$ and a user-defined target fidelity $\mathcal{L}_{target}$, we treat the gap $\mathcal{L}_{curr}- \mathcal{L}_{target}$ as an error budget that needs to be ``resolved'' by new primitives. The required primitive count $n_{add}$ is then predicted via a linear extrapolation of historical efficiency: $n_{add}= \left\lceil \frac{\mathcal{L}_{curr}- \mathcal{L}_{target}}{\Delta e_{add}}\right\rceil$.
In practice, we constrain $n_{add}\in [n_{min}, n_{max}]$ for stability. This feedback loop enables structural self-regulation: it aggressively adds primitives when the reconstruction gap is large and progressively restricts addition as fidelity improves, effectively balancing visual quality with primitive efficiency.

\subsubsection{(4) Optimization Objective.}
The optimization of each SVG layer is driven by a composite loss that balances reconstruction fidelity with geometric constraints. Let $\mathcal{S}_{k}$ denote the primitives for the current layer and $L_{k}$ be the target raster image with binary mask $M_{k}$. The total objective $\mathcal{L}$ is formulated as:
\vspace{-4pt}
\begin{equation}
    \label{eq:total_loss}\mathcal{L}= \mathcal{L}_{recon}+ \lambda_{mask}\mathcal{L}_{mask}
\end{equation}

The \textbf{reconstruction loss} $\mathcal{L}_{recon}= \|\mathcal{R}(\mathcal{S}_{k}) - L_{k}\|_{2}^{2}$ quantifies the pixel-wise discrepancy between the rendered image $\mathcal{R}(\mathcal{S}_{k})$ and the ground truth $L_{k}$. The \textbf{mask constraint loss} $\mathcal{L}_{mask}= \|\alpha(\mathcal{S}_{k}) \odot (\mathbf{1}- M_{k})\|_{2}^{2}$ enforces spatial boundaries by penalizing any rendered content outside the region defined by $M_{k}$, where $\alpha(\mathcal{S}_{k})$ represents the rendered alpha map. This constraint keeps the SVG primitives within the object's boundaries, preventing visual artifacts from spreading into the transparent background.

    \section{Experiments}
\label{sec:experiments} We evaluate our AmodalSVG against state-of-the-art vectorization methods in~\cref{subsec:comparisons} through quantitative comparisons on reconstruction fidelity and qualitative evaluation of layer decomposition quality. We then conduct ablation studies in~\cref{subsec:ablation} to analyze the contributions of components in Semantic Layer Peeling (SLP) and Adaptive Layered Vectorization (ALV). Finally,~\cref{subsec:applications} demonstrates the editing capabilities enabled by AmodalSVG.


\subsection{Comparisons}
\label{subsec:comparisons}


\subsubsection{Reconstruction Fidelity.}
Our comparisons are conducted on 30 raster images that span a wide range of real-world scenes. These images have varying aspect ratios, where the longest dimension ranges from 256 to 1024 pixels. For a fair comparison, we set an identical initial budget of 2,048 primitives for all methods that allow manual configuration. Furthermore, AdaVec~\cite{adavec_zhao_2025}, which is primarily designed for simple icons, utilizes its own initialization strategy and adaptive primitive counts, resulting in only about 200 primitives being used. LayerTracer~\cite{layertracer_song_2025} and LayerPeeler~\cite{layerpeeler_wu_2025} rely on external vectorization engines (\ie VTracer~\cite{vtracer_visioncortex_2020} and Recraft API~\cite{recraft_vectorizer_2024}); thus, the extracted number of primitives is not available.

\begin{table}[t]
    \centering
    \caption{\textbf{Quantitative comparison of SVG reconstruction fidelity.} The best results are \textbf{bolded} and the second best are \underline{underlined}.}
    \vspace{-1em}
    \label{tab:metrics} \resizebox{\linewidth}{!}{
    \begin{tabular}{l|cc|cccc|cc|c}
        \toprule \multirow{2}{*}[-0.25em]{Metric}                                     & \multicolumn{2}{c|}{Global Optimization} & \multicolumn{4}{c|}{Coarse-to-fine Layering} & \multicolumn{2}{c|}{Path-by-path Layering} & \multicolumn{1}{c}{Amodal Layering} \\
        \cmidrule(lr){2-3} \cmidrule(lr){4-7} \cmidrule(lr){8-9} \cmidrule(lr){10-10} & DiffVG~\cite{diffvg_li_2020}             & O\&R~\cite{optimizeandreduce_hirschorn_2024} & LIVE~\cite{live_ma_2022}                   & SGLIVE~\cite{sglive_zhou_2024}     & LayerTracer~\cite{layertracer_song_2025} & LayeredVec~\cite{layered_wang_2025} & LayerPeeler~\cite{layerpeeler_wu_2025} & AdaVec~\cite{adavec_zhao_2025} & \textbf{Ours}    \\
        \midrule MSE$\downarrow$                                                      & \underline{0.00098}                      & 0.00423                                      & 0.00110                                    & 0.01809                            & 0.00464                                  & 0.00099                             & 0.14324                                & 0.00891                        & \textbf{0.00072} \\
        PSNR$\uparrow$                                                                & \underline{30.52}                        & 23.76                                        & 30.07                                      & 18.00                              & 24.01                                    & 30.45                               & 9.72                                   & 20.61                          & \textbf{32.91}   \\
        SSIM$\uparrow$                                                                & 0.8860                                   & 0.7093                                       & 0.8863                                     & 0.6230                             & 0.7533                                   & \underline{0.8939}                  & 0.5591                                 & 0.6943                         & \textbf{0.9342}  \\
        LPIPS$\downarrow$                                                             & \underline{0.1075}                       & 0.3450                                       & 0.1635                                     & 0.3727                             & 0.2161                                   & 0.1307                              & 0.5211                                 & 0.3280                         & \textbf{0.0554}  \\
        \bottomrule
    \end{tabular} %
    }
    \vspace{-1em}
\end{table}

\begin{figure}[!t]
    \centering
    \includegraphics[width=1\linewidth]{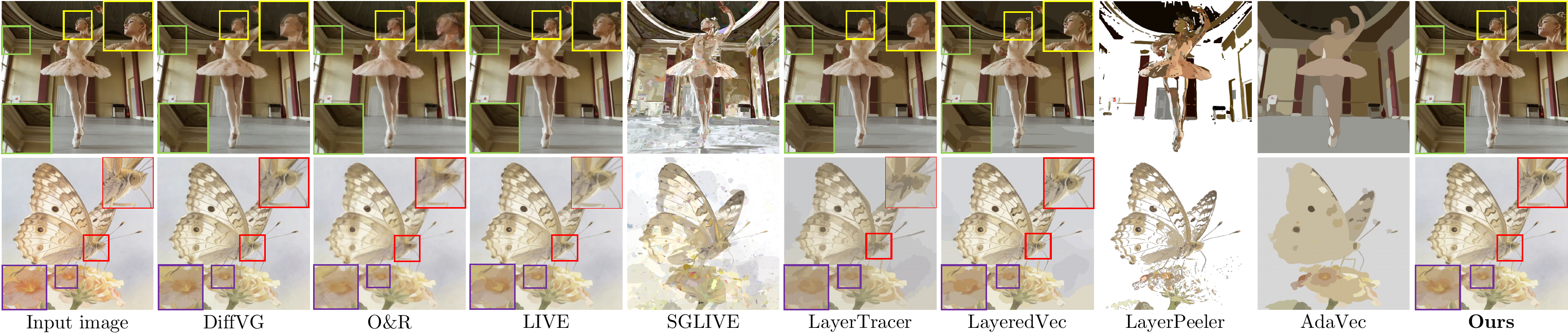}
    \vspace{-1.8em}
    \caption{\textbf{Qualitative comparison of SVG reconstruction fidelity.} We evaluate AmodalSVG against various state-of-the-art vectorization methods. See zoom-in regions (highlighted in boxes) for details. Our method achieves high reconstruction fidelity and preserves sharp edges and fine-grained textures than other methods.
    }
    \label{fig:compare_recon}
    \vspace{-1.8em}
\end{figure}

As shown in~\cref{fig:compare_recon}, our AmodalSVG achieves significantly sharper edges and more accurate textures compared to \textit{global optimization} schemes (\eg, DiffVG~\cite{diffvg_li_2020}) and \textit{coarse-to-fine layering} methods (\eg, LIVE~\cite{live_ma_2022}), which often suffer from severe blurriness and chaotic artifact accumulation. \textit{Path-by-path layering} methods (\eg, LayerPeeler~\cite{layerpeeler_wu_2025}) tend to oversimplify visual features, resulting in only coarse abstractions. Our method effectively preserves structurally sensitive regions, such as the ballerina's face and the butterfly's compound eye, which are typically lost or distorted in prior methods.

To quantitatively evaluate reconstruction fidelity, we employ four standard metrics: Mean Squared Error (MSE), Peak Signal-to-Noise Ratio (PSNR), Structural Similarity Index (SSIM)~\cite{ssim_wang_2004}, and Learned Perceptual Image Patch Similarity (LPIPS)~\cite{lpips_zhang_2018}. As shown in Tab.~\ref{tab:metrics}, our method consistently outperforms existing baselines across all quantitative metrics. Specifically, we achieve the lowest MSE (0.00072) and LPIPS (0.0554), indicating minimal pixel-wise discrepancies and superior perceptual quality. Our PSNR reaches 32.91 dB, significantly surpassing the second-best baseline DiffVG~\cite{diffvg_li_2020} (30.52 dB). This performance superiority can be attributed to our hierarchical decomposition paradigm. Benefiting from the semantically coherent segmentation provided by \textbf{SLP}, the optimization process is decoupled into individual layers, allowing the renderer to focus on reconstructing a single semantic entity at a time without interference from complex background textures. Furthermore, \textbf{ALV} enhances fidelity by dynamically inserting primitives to capture intricate local details while pruning redundant primitives to maintain overall efficiency.

\begin{figure}[!t]
    \centering
    \includegraphics[width=\linewidth]{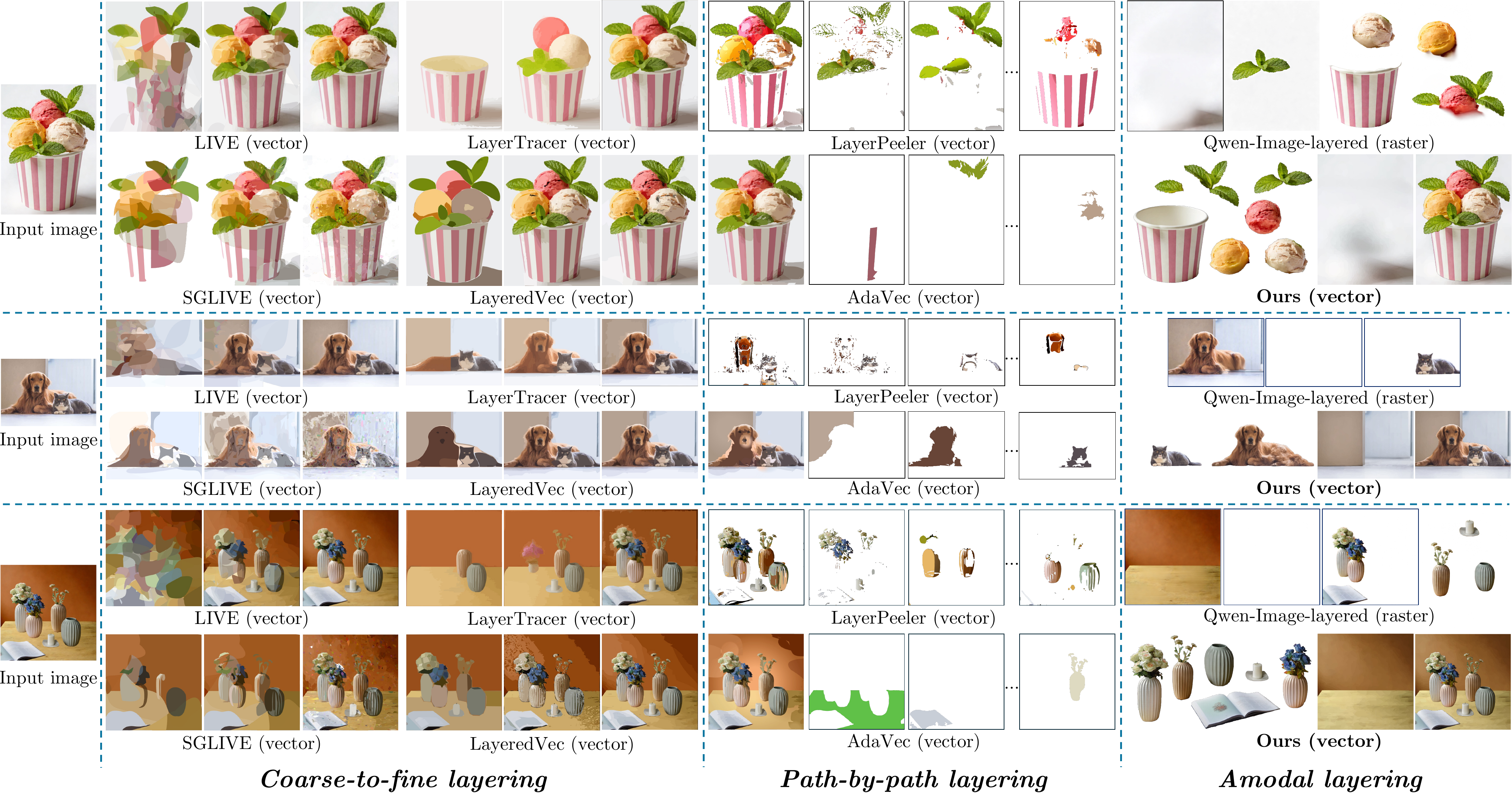}
    \vspace{-2em}
    \caption{\textbf{Qualitative comparison of SVG layering quality.} Compared with coarse-to-fine methods whose layers primarily serve the optimization process and often entangle multiple semantics and path-by-path approaches that generate fragmented segments, AmodalSVG produces structurally coherent layers. The generative Qwen-Image-Layered~\cite{qwenimagelayered_yin_2025} often entangles multiple semantics and produces empty layers in complex scenes, whereas our method accurately decouples semantic entities and reconstructs occluded regions.}
    \label{fig:compare_layer}
    \vspace{-1.5em}
\end{figure}
\subsubsection{Layering Quality.}
We evaluate the layer quality by visualizing the decomposed layers produced by our method and competing methods in~\cref{fig:compare_layer}. First, \textit{path-by-path layering} methods~\cite{layerpeeler_wu_2025, adavec_zhao_2025} often produce highly fragmented outputs where a single object is scattered across hundreds of separate layers, preventing any meaningful object-level editing, as illustrated in the ``Ice Cream'' (1st) and ``Still Life'' (3rd) cases. Second, in \textit{coarse-to-fine layering} methods \cite{live_ma_2022, sglive_zhou_2024, layertracer_song_2025, layered_wang_2025}, their layers serve to help the optimization process rather than representing meaningful objects. The resulting SVGs exhibit severe ``semantic entanglement'', where background primitives are mixed with foreground primitives, making it impossible to isolate the cup or the vases without breaking the visual integrity. Third, Qwen-Image-Layered~\cite{qwenimagelayered_yin_2025} attempts semantic separation at the pixel level, but it frequently struggle with the complex textures and occlusions found in real-world photography. This often leads to \textit{completely empty layers}, indicated by white placeholders in the ``Dog and Cat'' (2nd) and ``Still Life'' (3rd) cases, where critical scene components are lost.

In contrast, our layers show effective semantic decoupling. Besides, by restoring occluded regions through generative inpainting, \textbf{AmodalSVG} ensures that each layer is a structurally complete entity. For instance, as shown in~\cref{fig:compare_layer}, AmodalSVG can recover the cat's tail and the integral body of the dog behind the cat (2nd case), reconstruct the book in the bottom left corner and the desk surface and wall obscured by the vases (3rd case), and the background is also recovered as an integral entity. This enables precise per-object editing that is impossible with prior methods, as detailed in~\cref{subsec:applications}.

\begin{figure}[!t]
    \centering
    \vspace{-1em}
    \includegraphics[width=\linewidth]{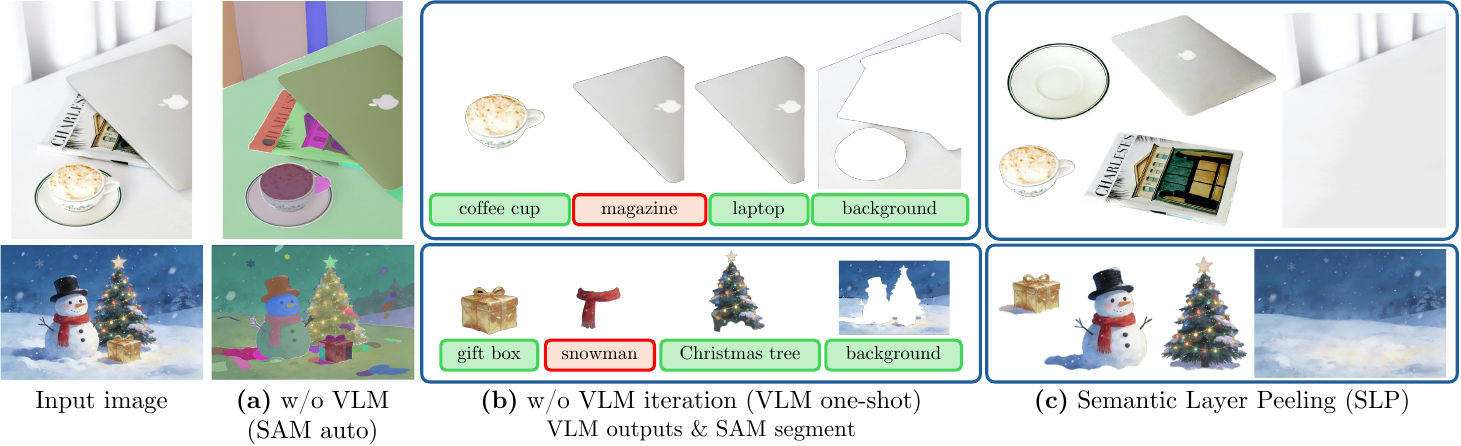}
    \vspace{-2em}
    \caption{\textbf{Ablation of Semantic Layer Peeling (SLP) strategies.} (a) \textbf{w/o VLM (SAM Auto):} Native SAM segmentation relies solely on low-level features, causing severe over-segmentation.
    (b) \textbf{w/o Iteration (One-shot):} Attempting to identify all objects simultaneously causes attention dispersion, leading to missed objects and imprecise boundary localization (object in red box). (c) \textbf{SLP (Ours):} Our iterative peel-and-inpaint strategy ensures robust entity extraction and structurally complete background recovery, yielding distinct and coherent semantic layers.}
    \label{fig:ablation_vlm}
    \vspace{-1.5em}
\end{figure}
\subsection{Ablation Study}
\label{subsec:ablation}
\subsubsection{Effectiveness of Semantic Layer Peeling.}
We validate the efficacy of our \textbf{Semantic Layer Peeling (SLP)} by comparing it against two baseline variants: \textit{Direct SAM Segmentation} (w/o VLM) and \textit{One-shot VLM Analysis} (w/o iteration). As shown in~\cref{fig:ablation_vlm}, the \textit{Direct SAM} baseline lacks high-level semantic understanding. Driven by local color and texture cues, it over-segments unified objects (\eg, the magazine) into patches (\cref{fig:ablation_vlm}(a)). The \textit{One-shot VLM} variant attempts to parse the entire scene in a single forward pass. However, dense prediction in complex scenes often overwhelms the VLM's attention mechanism, resulting in missed objects (\eg, the magazine) and vague semantic boundaries (\cref{fig:ablation_vlm}(b)). Furthermore, without iterative hybrid inpainting, this approach cannot reconstruct occluded regions, leaving empty spaces in the background. In contrast, our \textbf{SLP} strategy (\cref{fig:ablation_vlm}(c)) progressively decomposes the scene, creating layers that are semantically independent and structurally complete.

\begin{figure}[t]
    \begin{minipage}[t]{0.605\textwidth}
        \vspace{0pt}
        \centering
        \includegraphics[width=\linewidth]{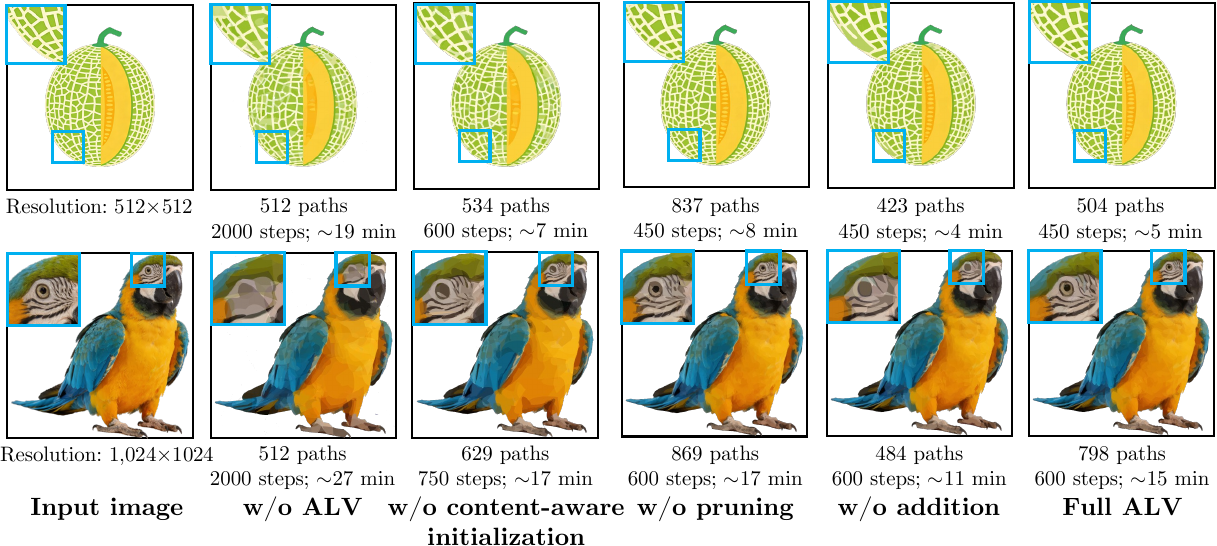}
        \vspace{-2em}
        \caption{\textbf{Qualitative ablation of ALV.} Content-aware initialization prevents convergence failures and facilitates structural recovery. Primitive addition helps capture high-frequency details (\eg, melon net), and primitive pruning removes redundant primitives for efficiency. Our full ALV achieves superior visual fidelity with an optimized primitive budget.}
        \label{fig:ablation_alv}
    \end{minipage}
    \hfill
    \begin{minipage}[t]{0.385\textwidth}
        \vspace{0pt}
        \centering
        \vspace{-0.75em}
        \captionof{table}{\textbf{Quantitative ablation of ALV components.} \textbf{Content-aware initialization} provides an informative prior, while \textbf{addition} recovers intricate details. Our \textbf{full ALV} balances visual fidelity and primitive efficiency, maintaining competitive metrics against the \textbf{w/o pruning} baseline without its redundant geometric complexity.} \label{tab:ablation_alv} \resizebox{\linewidth}{!}{
        \begin{tabular}{lccccc}
            \toprule Method          & \#Primitives & MSE$\downarrow$     & PSNR$\uparrow$    & SSIM$\uparrow$     & LPIPS$\downarrow$  \\
            \midrule w/o ALV         & 512     & 0.00279             & 26.08             & 0.8620             & 0.1106             \\
            w/o content-aware init   & 567     & 0.00232             & 27.25             & 0.8832             & 0.0911             \\
            w/o primitive pruning         & 897     & \textbf{0.00067}    & \textbf{31.75}    & \textbf{0.9342}    & \textbf{0.0386}    \\
            w/o primitive addition        & 438     & 0.00128             & 29.30             & 0.9186             & 0.0516             \\
            \rowcolor{gray!15} \textbf{Full ALV (Ours)} & 683     & \underline{0.00071} & \underline{31.48} & \underline{0.9336} & \underline{0.0392} \\
            \bottomrule
        \end{tabular}
        }
    \end{minipage}
    \vspace{-1em}
\end{figure}
\subsubsection{Effectiveness of Adaptive Layered Vectorization (ALV).}
We evaluate the individual contributions of ALV's core components in~\cref{fig:ablation_alv} and \cref{tab:ablation_alv} using a representative subset of the test dataset. The \textbf{content-aware initialization} provides an essential prior. Without it, randomly initialized primitives often scatter outside the active mask, which increases computational overhead without improving visual quality and causes significant structural misalignment (\eg, PSNR drops from 31.48~dB to 27.25~dB despite more optimization steps). Furthermore, \textbf{primitive addition} is essential for capturing high-frequency fine details. Omitting it causes a failure in recovering textures like the melon's surface net or the fine feathers around the parrot's eye, resulting in a substantial performance drop (PSNR drops to 29.30~dB and LPIPS rises to 0.0516). Conversely, \textbf{primitive pruning} enforces efficiency by eliminating visually redundant primitives. Although the ``w/o pruning'' baseline yields marginally higher numerical metrics in \cref{tab:ablation_alv}, it results in excessive geometric complexity (\eg, 869 \vs 798 primitives for the parrot) with negligible perceptual gain. Ultimately, our \textbf{full ALV} framework achieves a superior trade-off, delivering high-fidelity reconstruction through a compact and efficiently distributed primitive budget.

\begin{table}[t]
    \centering
    \caption{\textbf{Comparison of editing capabilities across different vectorization paradigms.} \cmark~denotes full support, \xmark~denotes no support, and \pmark~indicates limited or conditional support (\eg, only for simple icons or resulting in visual artifacts/holes).}
    \vspace{-1em}
    \label{tab:capability_comparison} \resizebox{\linewidth}{!}{
    \begin{tabular}{lcccccccc}
        \toprule Paradigm                                                                                                                                         & \multicolumn{1}{c}{\begin{tabular}[c]{@{}c@{}}Semantic\\ Decoupling\end{tabular}} & \multicolumn{1}{c}{\begin{tabular}[c]{@{}c@{}}Structural\\ Completion\end{tabular}} & Recolor & Resize & Reposition & Reorder & Remove & Replace \\
        \midrule Global Optimization~\cite{potrace_selinger_2003, vtracer_visioncortex_2020, diffvg_li_2020, im2vec_reddy_2021, optimizeandreduce_hirschorn_2024} & \xmark                                                                            & \xmark                                                                              & \xmark  & \xmark & \xmark     & \xmark  & \xmark & \xmark  \\
        Path-by-path Layering~\cite{adavec_zhao_2025, layerpeeler_wu_2025}                                                                                        & \xmark                                                                            & \pmark                                                                              & \pmark  & \xmark & \xmark     & \xmark  & \xmark & \xmark  \\
        Coarse-to-Fine Layering~\cite{live_ma_2022, sglive_zhou_2024, layertracer_song_2025, layered_wang_2025}                                                   & \pmark                                                                            & \xmark                                                                              & \pmark  & \pmark & \pmark     & \xmark  & \pmark & \pmark  \\
        \midrule \textbf{Amodal Layering (Ours)}                                                                                                                  & \cmark                                                                            & \cmark                                                                              & \cmark  & \cmark & \cmark     & \cmark  & \cmark & \cmark  \\
        \bottomrule
    \end{tabular}
    }
    \vspace{-2em}
\end{table}
\begin{figure}[!t]
    \centering
    \vspace{-1.2em}
    \includegraphics[width=0.93\linewidth]{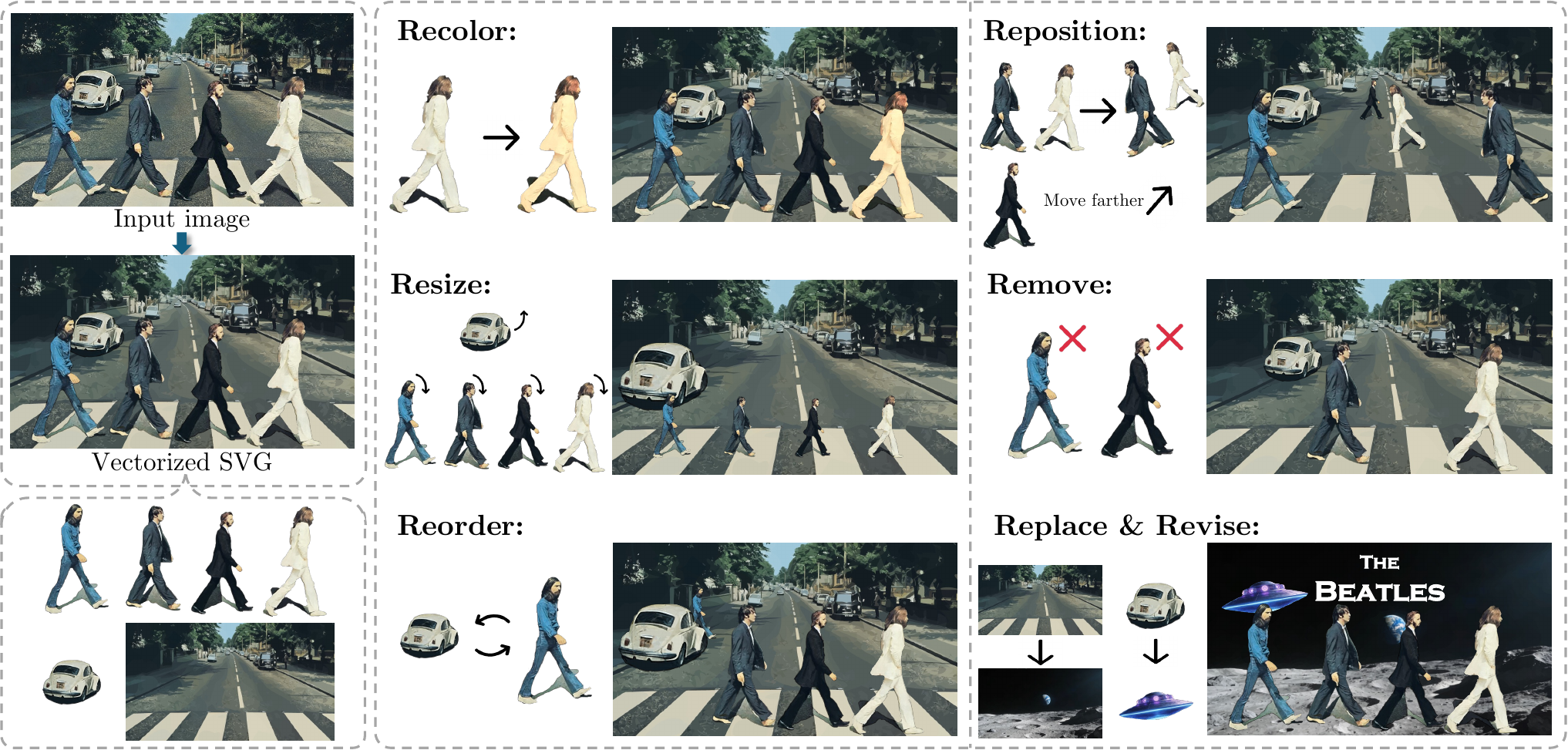}
    \vspace{-1em}
    \caption{\textbf{Structural editability and downstream applications.} With its structurally complete and semantically decoupled layers, \textbf{AmodalSVG} enables seamless object-level manipulations and scene modifications without background gaps.}
    \label{fig:application}
    \vspace{-2em}
\end{figure}
\subsection{Applications}
\label{subsec:applications} Beyond high-fidelity reconstruction from the raster domain to vector domain, the primary utility of \textbf{AmodalSVG} lies in its capacity to generate \textit{structurally complete} and \textit{semantically decoupled} vector layers. Table~\ref{tab:capability_comparison} compares editing capabilities across different vectorization paradigms. By restoring occluded regions through SLP, our framework enables precise object-level editing that is impossible with modal vectorization. As demonstrated in~\cref{fig:teaser} and~\cref{fig:application}, individual semantic layers can be seamlessly \textit{recolored}, \textit{repositioned}, \textit{reordered}, \textit{resized}, \textit{removed}, or \textit{replaced} without introducing visual artifacts or background gaps. 

Furthermore, our representation simplifies overall scene editing, such as adding new text elements or entirely swapping the background contexts. This flexibility validates \textbf{AmodalSVG} as a robust foundation for professional graphic design and iterative creative workflows.

    \section{Conclusion}
\label{sec:conclusion}
In this paper, we introduced a new task, \textbf{Amodal Image Vectorization}, which requires parsing raster images into semantically independent and structurally complete vector layers, including recovering occluded regions. To address this, we presented a novel framework, \textbf{AmodalSVG}. Unlike existing methods producing unstructured or semantically entangled primitives, AmodalSVG leverages the semantic reasoning of VLMs to progressively decompose scenes into independent, amodally complete raster layers prior to vectorization. Our \textbf{Semantic Layer Peeling (SLP)} strategy effectively handles complex occlusions through hybrid inpainting, while \textbf{Adaptive Layered Vectorization (ALV)} dynamically manages primitive complexity, ensuring high visual fidelity with a compact, optimized primitive budget for each layer. Extensive experiments demonstrate that our approach not only surpasses state-of-the-art methods in reconstruction quality but also unlocks practical editing capabilities, such as object-level reordering and overall scene editing, which were previously impossible.
\paragraph{Limitations and Future Work.}
Despite its promising results, AmodalSVG relies on pre-trained vision-language models (VLMs) to extract each layer in the raster domain. As a result, errors in object identification or missed segmentation of secondary effects, such as mirror-like reflections from reflective surfaces, may degrade the quality of the generated layers. However, we believe that as the power of VLMs improves, so will our framework. Furthermore, the iterative peeling incurs considerable computational overhead. Future work will focus on accelerating the pipeline, expanding the representation to support global lighting effects, and improving the robustness of layer separation for complex topologies.

    %
    %
    \bibliographystyle{splncs04}
    \bibliography{main}

    \renewcommand{\thefigure}{S\arabic{figure}}
\setcounter{figure}{0}
\renewcommand{\thetable}{S\arabic{table}}
\setcounter{table}{0}
\renewcommand{\theequation}{S\arabic{equation}}
\setcounter{equation}{0}

\newpage
\appendix

\begin{center}
    \textbf{\Large Supplementary Material}
\end{center}

\section*{Overview}
In this supplementary material, we provide additional details and discussions related to our work on \textbf{AmodalSVG}. Specifically, this document covers the following aspects:
\begin{itemize}[left=0pt]
    \item \textbf{Implementation Details} (\cref{sec:supp_impl_details}): We provide comprehensive hyperparameters and configuration settings for both the Semantic Layer Peeling (SLP) and Adaptive Layered Vectorization (ALV) stages to ensure reproducibility.

    \item \textbf{Prompts for Semantic Layer Peeling} (\cref{sec:supp_prompts}): We present the exact instruction templates and chain-of-thought prompts utilized by the Vision-Language Model during the iterative peeling process.

    \item \textbf{Additional Ablation Studies} (\cref{sec:supp_additional_ablations}): We provide additional ablation studies to justify our design choices, including the hybrid inpainting strategy in SLP, as well as the primitive pruning and addition strategies in ALV.

    \item \textbf{Failure Cases} (\cref{sec:supp_failure_cases}): We discuss the limitations of our current framework, such as errors in complex object identification and missed segmentation of secondary visual effects like reflections.
\end{itemize}

\begin{figure}[!t]
    \centering
    \begin{tcolorbox}
        [colback=MainPurple!5!white, colframe=MainPurple!70!black, title=System Prompt for Semantic Layer Peeling, fonttitle=\bfseries, left=5pt, right=5pt, top=5pt, bottom=5pt, sharp corners, boxrule=0.5pt ] \ttfamily\small Identify ONE SINGLE foreground object instance (not all of them).

        \textbf{What counts as BACKGROUND (return is\_background\_only=true):}
        \begin{itemize}[left=0pt, itemsep=0pt, topsep=2pt, parsep=0pt]
            \setlength{\parskip}{0pt}

            \item floor, ground, wall, ceiling, sky, road, grass, water, sand, snow

            \item building exterior, room interior, landscape, scenery

            \item ANY scene element without a clear subject
        \end{itemize}

        \textbf{What counts as FOREGROUND:}
        \begin{itemize}[left=0pt, itemsep=0pt, topsep=2pt, parsep=0pt]
            \setlength{\parskip}{0pt}

            \item person, animal, vehicle, furniture, item, food, plant (potted)

            \item Something you can PICK UP or MOVE
        \end{itemize}

        \textbf{Step 1: Check occlusion}
        \begin{itemize}[left=0pt, itemsep=0pt, topsep=2pt, parsep=0pt]
            \item Which object is IN FRONT? (not blocked)

            \item The frontmost complete object = target
        \end{itemize}

        \textbf{Step 2: If multiple same objects (\eg, 3 dogs)}
        \begin{itemize}[left=0pt, itemsep=0pt, topsep=2pt, parsep=0pt]
            \item Pick ONLY ONE - the most complete/frontmost

            \item Add position or color: "left dog", "center dog", "brown dog"
        \end{itemize}

        \textbf{Step 3: Check for shadows}
        \begin{itemize}[left=0pt, itemsep=0pt, topsep=2pt, parsep=0pt]
            \item Does the object cast a visible shadow on the surface?

            \item Shadow = darker area on floor/table caused by object blocking light

            \item Do NOT confuse shadow with: dark-colored objects, reflections, or dark textures
        \end{itemize}

        Return JSON.

        Requirements for the output:
        \begin{itemize}[left=0pt, itemsep=0pt, topsep=2pt, parsep=0pt]
            \item `description' should be concise but specific enough to uniquely identify the target object

            \item `keywords' should contain a few useful segmentation terms for the same object
        \end{itemize}
        \begin{verbatim}
{
  "is_background_only": false,
  "description": "A grey cat",
  "keywords": ["cat"],
  "has_shadow": true
}
\end{verbatim}

        If only background/scene: \begin{verbatim}
{
  "is_background_only": true,
  "description": "",
  "keywords": [],
  "has_shadow": false
}
\end{verbatim}
    \end{tcolorbox}
    \vspace{-1.5em}
    \caption{\textbf{System Prompt used for Semantic Layer Peeling.} This structured prompt guides the VLM to sequentially reason about occlusion relationships, isolate the frontmost semantic entity, and format its output as an easily parsable JSON tuple.}
    \label{fig:supp_prompt}
    \vspace{-1.5em}
\end{figure}

\section{Implementation Details}
\label{sec:supp_impl_details} Our \textbf{AmodalSVG} framework is implemented in PyTorch using DiffVG~\cite{diffvg_li_2020} for differentiable rasterization on a single NVIDIA A100 GPU.

\paragraph{Semantic Layer Peeling (SLP).}
We employ Qwen3-VL-8B-Instruct~\cite{qwen3vl_bai_2025} with a maximum of 10 peeling iterations. For Grounded Segmentation~\cite{groundedsam_ren_2024}, we combine Grounding DINO~\cite{groundingdino_liu_2024} for bounding box prediction and SAM ViT-H~\cite{sam_kirillov_2023} for mask generation. For Background Restoration, we sequentially apply LaMa~\cite{lama_suvorov_2022} and Flux.1 Fill [DEV]~\cite{flux_fill_blackforestlab_2024} (28 denoising steps, guidance scale 10.0).

\paragraph{Adaptive Layered Vectorization (ALV).}
We set the initial primitive budget $N_{0}= 512$, distributed proportionally to the layer's pixel area. Primitive addition and pruning activate at iteration 200 with an evaluation interval of 100 steps. The residual error field $E(p)$ uses \texttt{grad-l2} maps ($T = 0.5$). For the error budget estimation, we use a sliding window $N = 3$, marginal contribution $\Delta e_{add}= 10^{-3}$, target fidelity $\mathcal{L}_{target}= 1\times 10^{-3}$, and clamp the predicted primitive count $n_{add}\in [5, 100]$. The pruning threshold is $\tau_{p}= 10$. Each layer is optimized for 2,000 iterations using the Adam optimizer (learning rates: 1.0 for points, 0.01 for colors) with a linear decay schedule (ratio 0.4) and mask constraint weight $\lambda_{mask}= 0.5$.

\section{Prompts for Semantic Layer Peeling}
\label{sec:supp_prompts} In this section, we detail the text prompts used by the Vision-Language Model (VLM, \ie, Qwen3-VL~\cite{qwen3vl_bai_2025}) to guide the Semantic Layer Peeling (SLP) process. Our SLP framework relies on the VLM to perform complex scene parsing, specifically requiring it to reason about depth and occlusion relationships to identify the foremost foreground entity at each iteration. As detailed in~\cref{fig:supp_prompt}, we design a Chain-of-Thought (CoT) prompt template that instructs the model to carefully analyze spatial overlaps before making a conclusion. Once the VLM confirms the existence of a distinct foreground object, it outputs a semantic tuple containing a dense visual description and discriminative keywords for the subsequent GroundedSAM~\cite{groundedsam_ren_2024} segmentation. If only the background context remains, it outputs a termination signal.

\section{Additional Ablation Studies}
\label{sec:supp_additional_ablations}
\begin{figure}[!t]
    \centering
    \includegraphics[width=\linewidth]{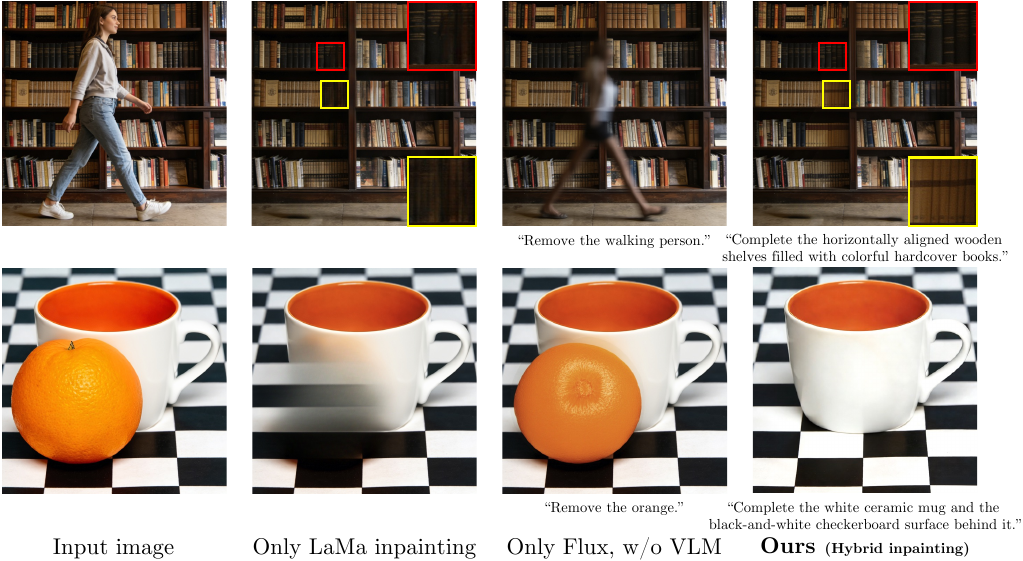}
    \vspace{-2em}
    \caption{\textbf{Ablation of the hybrid inpainting strategy.} We evaluate our synergistic inpainting pipeline against single-model baselines. While \textit{Only LaMa} maintains structural continuity but lacks high-frequency detail (resulting in blurry textures or streaky artifacts), \textit{Only Flux (w/o VLM guidance)} often fails to accurately localize and remove foreground entities. Our \textbf{Hybrid inpainting} strategy, guided by VLM semantic parsing, successfully reconstitutes complex backgrounds (\eg, bookshelf textures and cup geometry) with high fidelity, ensuring structural integrity for subsequent vectorization.}
    \label{fig:supp_ablation_inpainting}
    \vspace{-1.5em}
\end{figure}
\subsection{Comparison of Inpainting Strategies}
To ensure each peeled layer remains structurally complete, we propose a synergistic inpainting pipeline comprising LaMa, VLM guidance, and FLUX. We justify this design by comparing it against alternative fundamental baselines, with visual results provided in~\cref{fig:supp_ablation_inpainting}.

\paragraph{LaMa Only.}
While LaMa~\cite{lama_suvorov_2022} is efficient at synthesizing stationary background textures (\eg, walls or plain surfaces), it lacks the generative capacity for intricate structured objects. When hallucinating a complex underlying layer over a large occluded region, it frequently fails to reconstruct fine-grained semantic details. Consequently, the inpainted region often manifests as a blurry, disconnected color smudge that loses the crisp boundaries of the original objects.

\paragraph{FLUX Only (w/o VLM Guidance).}
In this baseline, we directly feed the masked image into the native FLUX~\cite{flux_fill_blackforestlab_2024} inpainting pipeline, bypassing our proposed cascade. Without an initial visible geometry, the masked region is a complete void, meaning a VLM has no visual context to infer the missing semantics. Consequently, FLUX is forced to bridge massive holes blindly. To guide the diffusion model without VLM's affirmative analysis, one typically resorts to hard-coded negative prompts (\eg, ``an image \textit{without} [Foreground Object]''). However, this lacks explicit positive guidance, causing severe structural discontinuities (\eg, broken perspective lines or misaligned repeating patterns). Furthermore, negative prompting inevitably triggers the ``pink elephant'' effect, stubbornly regenerating distorted ghosts and remnants of the foreground entity.

\paragraph{Ours (LaMa + VLM + FLUX).}
Our pipeline logically cascades these components to create a closed loop of structural and semantic guidance. First, LaMa effectively connects the foundational geometry across the wide mask, ensuring geometric alignment. Crucially, this intermediate result—although blurry—reveals enough structural context for the VLM to visually recognize the occluded semantics. The VLM then analyzes this LaMa output and issues a precise, \textit{positive} prompt describing the exact missing background components. Finally, driven by this accurate, explicitly affirmative prompt and initialized by LaMa's structural prior, FLUX safely renders high-fidelity, immaculate textures consistent with the background context.

\subsection{Comparison of Primitive Pruning Strategies}
In our Adaptive Layered Vectorization (ALV) stage, dynamically pruning redundant primitives is essential for maintaining a compact and editable SVG topology without sacrificing visual fidelity. To justify our occlusion-aware pruning formulation, we compare it against typical heuristic baselines and an ``Oracle'' brute-force upper bound. We evaluate the reconstruction quality across four metrics: MSE ($\downarrow$), PSNR ($\uparrow$), SSIM~\cite{ssim_wang_2004} ($\uparrow$), and LPIPS~\cite{lpips_zhang_2018} ($\downarrow$), at three progressive pruning ratios ($10\%$, $20\%$, and $30\%$ primitive reduction). To provide a standardized baseline and isolate the impact of different pruning operations, all 10 original SVGs used in this experiment are uniformly generated using DiffVG~\cite{diffvg_li_2020} with a fixed budget of 2,048 primitives. Note that the ``pruning ratio'' reflects the proportion of dropped primitives relative to this initial set before further pruning. While our framework practically uses an absolute path-area threshold $\tau_{p}$ for automated pruning, we convert this continuous score ($\mathcal{C}_{i}$) into strict percentile rankings for this ablative evaluation to ensure identical removal quantities across all competing methods.

\begin{table}[t]
    \centering
    \small
    \setlength{\tabcolsep}{4.5pt}
    \caption{\textbf{Quantitative comparison of primitive pruning strategies.} We evaluate different pruning strategies under three reduction ratios ($10\%$, $20\%$, and $30\%$) and report the average performance over 10 SVGs. For metrics (MSE, PSNR, SSIM, and LPIPS) across different reduction ratios, the theoretical upper bound (\textit{Oracle}) is shown in \textbf{bold}, and the best practical performance is \underline{underlined}. Runtime is averaged per SVG. Our method significantly outperforms naive Area-based or Opacity-based heuristics, achieving comparable fidelity to the brute-force Oracle bound while offering a massive $\sim20\times$ speedup.}
    \vspace{-1em}
    \label{tab:supp_pruning_quant} \resizebox{\textwidth}{!}{%
    \begin{tabular}{llccccc}
        \toprule \multirow{2}{*}{\textbf{Prune ratio}} & \multirow{2}{*}{\textbf{Method}} & \multicolumn{4}{c}{\textbf{Quality}} & \multirow{2}{*}{\textbf{Runtime (s) $\downarrow$}} \\
        \cmidrule(lr){3-6}                             &                                  & \textbf{MSE $\downarrow$}            & \textbf{PSNR $\uparrow$}                          & \textbf{SSIM $\uparrow$} & \textbf{LPIPS $\downarrow$} &                   \\
        \midrule                                       & Area                             & $1.61\times10^{-4}$                  & 40.24                                             & 0.9896                   & 0.0104                      & \underline{0.533} \\
                                                       & Opacity                          & $2.41\times10^{-4}$                  & 37.76                                             & 0.9856                   & 0.0177                      & \textbf{0.244}    \\
                                                       & Oracle                           & $\mathbf{5.43\times10^{-6}}$         & $\mathbf{\infty}$                                 & \textbf{0.9997}          & \textbf{0.0020}             & 129.049           \\
        \rowcolor{gray!15} \multirow{-4}{*}{10\%}      & Ours                             & $\underline{4.75\times10^{-5}}$      & \underline{52.49}                                 & \underline{0.9946}       & \underline{0.0065}          & 6.468             \\
        \midrule                                       & Area                             & $3.83\times10^{-4}$                  & 36.57                                             & 0.9769                   & 0.0232                      & \underline{0.322} \\
                                                       & Opacity                          & $7.23\times10^{-4}$                  & 32.95                                             & 0.9647                   & 0.0400                      & \textbf{0.247}    \\
                                                       & Oracle                           & $\mathbf{2.12\times10^{-5}}$         & \textbf{50.48}                                    & \textbf{0.9965}          & \textbf{0.0071}             & 129.047           \\
        \rowcolor{gray!15} \multirow{-4}{*}{20\%}      & Ours                             & $\underline{1.52\times10^{-4}}$      & \underline{40.65}                                 & \underline{0.9868}       & \underline{0.0168}          & 6.469             \\
        \midrule                                       & Area                             & $6.04\times10^{-4}$                  & 34.18                                             & 0.9651                   & 0.0377                      & \underline{0.327} \\
                                                       & Opacity                          & 0.0013                               & 30.38                                             & 0.9438                   & 0.0657                      & \textbf{0.265}    \\
                                                       & Oracle                           & $\mathbf{5.66\times10^{-5}}$         & \textbf{45.08}                                    & \textbf{0.9913}          & \textbf{0.0154}             & 129.052           \\
        \rowcolor{gray!15} \multirow{-4}{*}{30\%}      & Ours                             & $\underline{3.42\times10^{-4}}$      & \underline{36.32}                                 & \underline{0.9760}       & \underline{0.0332}          & 6.485             \\
        \bottomrule
    \end{tabular}
    }
    \vspace{-1.5em}
\end{table}
\paragraph{Baselines.}
To comprehensively benchmark our approach, we establish the following representative heuristic strategies (frequently utilized in contemporary area- or opacity-based pruning pipelines) for comparison:
\begin{itemize}[left=0pt]
    \item \textbf{Area-based Pruning:} Iteratively removes primitives with the smallest spatial area, under the assumption that tiny strokes contribute marginally to human perception.

    \item \textbf{Opacity-based Pruning:} Removes primitives with the lowest alpha channel values (most transparent), assuming they carry the least color density.

    \item \textbf{Oracle (Ground Truth Bound):} A computationally expensive strategy where we tentatively drop each primitive one-by-one, recalculate the true global loss difference, and permanently prune the one causing the least degradation. This acts as our theoretical upper bound.

    \item \textbf{Ours (Occlusion-aware Pruning):} Our proposed method evaluates the effective visual contribution score ($\mathcal{C}_{i}$) of primitives by computing their cumulative occlusion from overlying layers, preserving structural paths over flat redundancy.
\end{itemize}
\paragraph{Quantitative Analysis.}
As shown in~\cref{tab:supp_pruning_quant}, pruning by strict heuristics quickly deteriorates visual fidelity. The Area-based method rapidly increases MSE at higher fractions (\eg, 30\%), as it tends to eliminate small but critical high-frequency primitives (\eg, sharp object edges or dense textures). The Opacity-based baseline severely compromises LPIPS~\cite{lpips_zhang_2018} and SSIM~\cite{ssim_wang_2004} scores, since semi-transparent primitives often play crucial roles in blending soft shadows and simulating lighting transitions. In contrast, while the exhaustive Oracle strategy provides the theoretical upper bound for rendering quality, it requires an intractable computational cost ($>380$ seconds per SVG). Our occlusion-aware strategy achieves highly competitive metrics—consistently ranking second only to the Oracle bound across all quality indicators and progressive pruning ratios. More importantly, by leveraging the cumulative occlusion proxy, our method achieves a massive $\sim20\times$ speedup (18.69s \vs 386.43s) over the brute-force Oracle approach. This demonstrates that our method effectively bypasses the $\mathcal{O}(N^{2})$ computational bottleneck while maintaining superior structural integrity compared to naive heuristics.

\begin{figure}[!t]
    \centering
    \includegraphics[width=\linewidth]{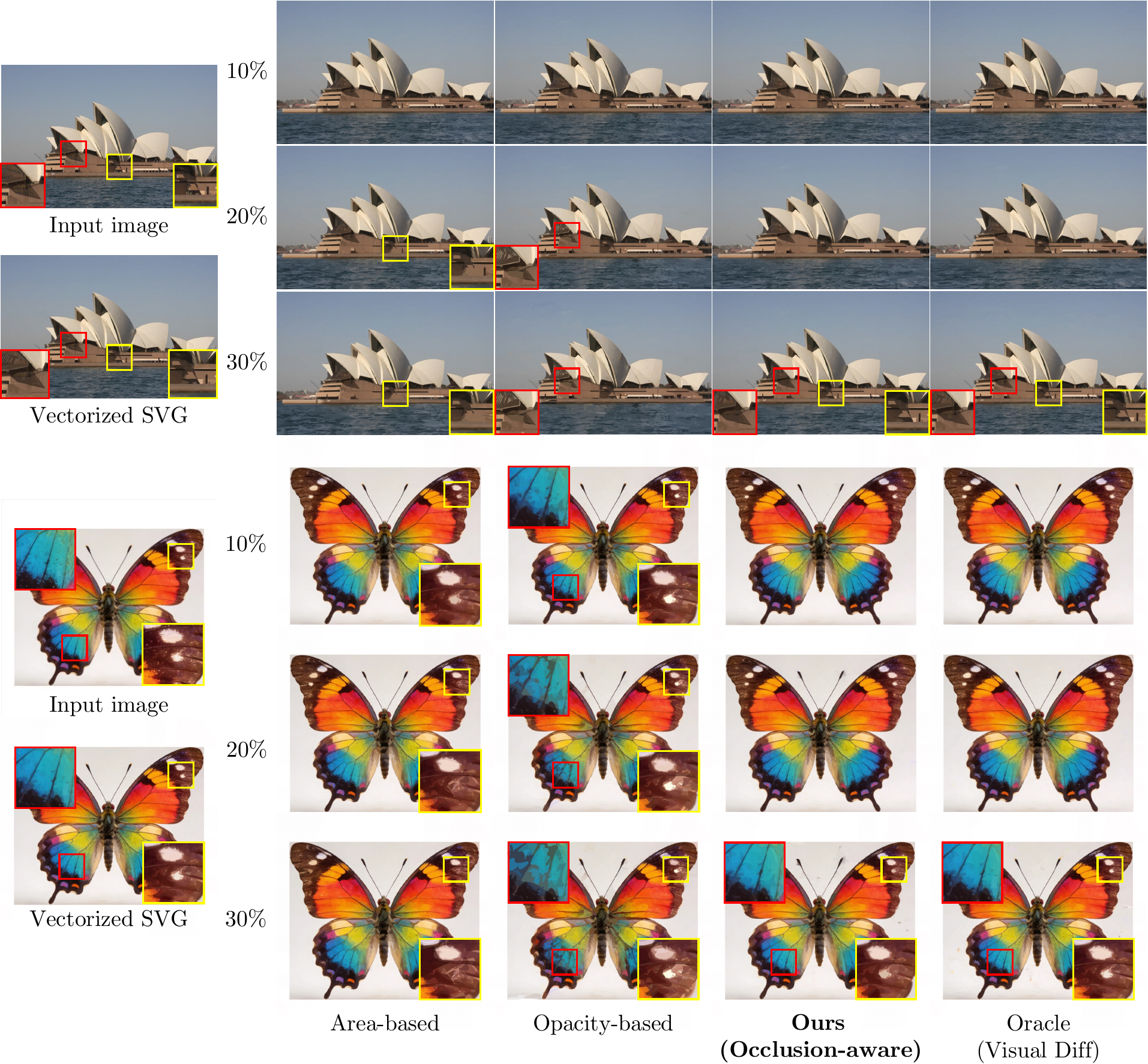}
    \vspace{-2em}
    \caption{\textbf{Qualitative comparison of primitive pruning strategies.} We evaluate our \textit{Occlusion-aware} strategy against \textit{Area-based} and \textit{Opacity-based} heuristics, with \textit{Oracle} serving as the theoretical ground truth upper bound. At progressive pruning ratios (10\%--30\%), the area-based method tends to eliminate fine structural details (\textit{e.g.}, sharp edges of the Opera House), while the opacity-based method disrupts complex color blending and gradients (\textit{e.g.}, the butterfly's wing patterns). In contrast, our method identifies redundant primitives within the occlusion hierarchy, maintaining visual fidelity almost identical to the exhaustive Oracle bound.}
    \label{fig:supp_pruning_qual}
    \vspace{-1.5em}
\end{figure}
\paragraph{Qualitative Analysis.}
To further validate the effectiveness of our occlusion-aware proxy, we compare it against common heuristics in~\cref{fig:supp_pruning_qual}. We visualize the results across progressive pruning budgets, ranging from 10\% to 30\% reduction of the initial primitive count.

As illustrated in the zoom-in regions, \textbf{Area-based pruning} proves inadequate for preserving structural features. Since high-frequency details and sharp edges are often represented by small-scale primitives, a purely area-driven approach prematurely eliminates these critical components, leading to overly simplified geometry in scenes like the ``Sydney Opera House''. Similarly, \textbf{Opacity-based pruning} fails to account for the composite visual effects generated by stacked semi-transparent primitives. In the ``Butterfly'' case, removing low-opacity primitives results in catastrophic color shifts and the destruction of subtle gradients, as these transparent elements are vital for the final color synthesis.

Our \textbf{Occlusion-aware} strategy, by contrast, evaluates visual importance within the global rendering occlusion hierarchy. It effectively identifies and drops primitives that are significantly obscured or contribute minimally to the observable alpha area. Consequently, even at an aggressive pruning ratio of 30\%, our method prevents the premature collapse of shape geometry and color dynamics. It reliably drops redundant filler strokes embedded within flat color regions, maintaining crisp semantic boundaries and soft structural lighting effects that are visually indistinguishable from the theoretical \textbf{Oracle} bound.

\subsection{Effect of the Temperature Hyperparameter $T$}
\begin{figure}[!t]
    \centering
    \includegraphics[width=\linewidth]{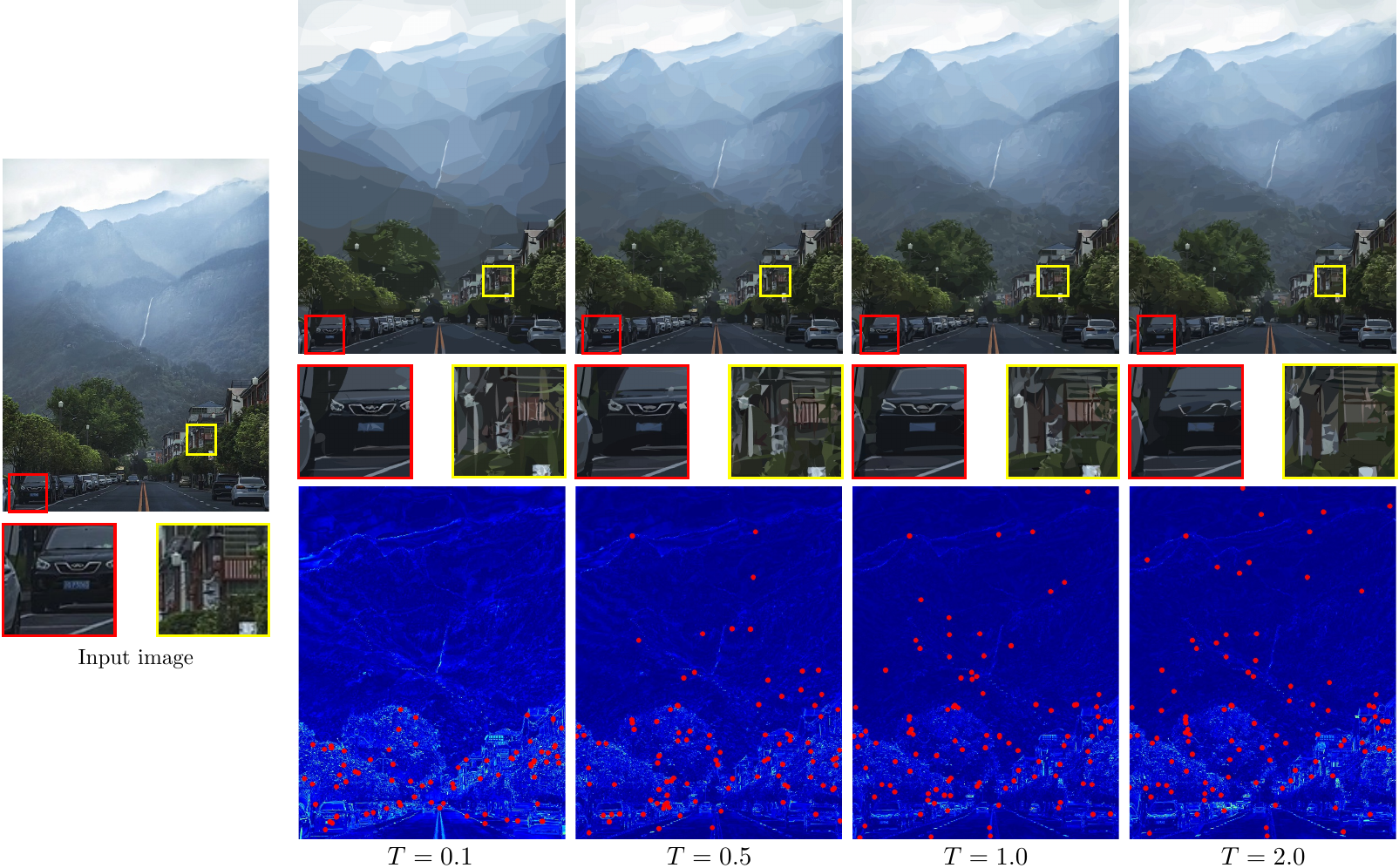}
    \vspace{-2em}
    \caption{\textbf{Effect of the temperature hyperparameter $T$ on primitive addition.} We visualize the final reconstruction results (top), detailed zoom-in regions (middle), and the corresponding primitive sampling distributions (bottom, where red dots denote newly sampled primitives over the residual error map). At \textbf{low temperature} ($T=0.1$), the distribution sharpens, causing primitives to greedily concentrate on high-frequency error peaks (\eg, object contours), leaving smoother gradient regions structurally under-fitted. At \textbf{high temperature} ($T=2.0$), the distribution flattens into nearly uniform exploration, overlooking localized structural details. This results in blurred textures and collapsed geometry, as evidenced in the car (red box) and architecture (yellow box). Our default intermediate value (\textbf{$T=0.5$}) achieves an optimal balance, ensuring both crisp structural boundaries and regional color consistency.}
    \label{fig:supp_ablation_temperature}
    \vspace{-1.5em}
\end{figure}

In our \textit{Importance-driven Primitive Addition} module, the temperature hyperparameter $T$ modulates the spatial distribution when sampling new primitives from the residual error field $E(p)$. This mechanism is formulated via a temperature-scaled weighting: $\mathcal{P}(p) \propto E(p)^{1/T}$. As analyzed below, $T$ governs the \textit{exploration-exploitation} trade-off during primitive allocation:

\begin{itemize}[left=0pt]
    \item \textbf{Exploitation-focused} ($T \to 0$): The probability distribution $\mathcal{P}(p)$ sharpens significantly, heavily dominating the peak values of the error map. As illustrated in the bottom row of \cref{fig:supp_ablation_temperature} ($T=0.1$), this enforces strict \textit{exploitation}, where new primitives (red dots) densely cluster exactly on high-contrast object boundaries. While this rapidly corrects salient structural errors, it causes localized geometric over-parameterization and leaves broader, mid-frequency regions (\eg, the background mountains) entirely unrefined.

    \item \textbf{Exploration-focused} ($T \gg 1$): The distribution $\mathcal{P}(p)$ flattens toward a uniform state, encouraging \textit{exploration}. This causes new primitives to scatter broadly across the canvas, regardless of local error magnitudes. As shown in the $T=2.0$ case, this uniform-like sampling allocates insufficient primitive density to structurally intricate zones. Consequently, sharp features like the car's headlights and the architecture's facade become blurred and over-simplified, as demonstrated in the zoom-in regions.
\end{itemize}

Empirically, we adopt $T = 0.5$ as our default setting. As substantiated in \cref{fig:supp_ablation_temperature}, this value yields an optimal balance: it ensures high primitive density along high-frequency contours to maintain structural sharpness, while appropriately distributing enough primitives across internal regions to resolve gradient discrepancies and maintain global color homogeneity.

\section{Failure Cases}
\label{sec:supp_failure_cases}
\begin{figure}[!t]
    \centering
    \includegraphics[width=0.8\linewidth]{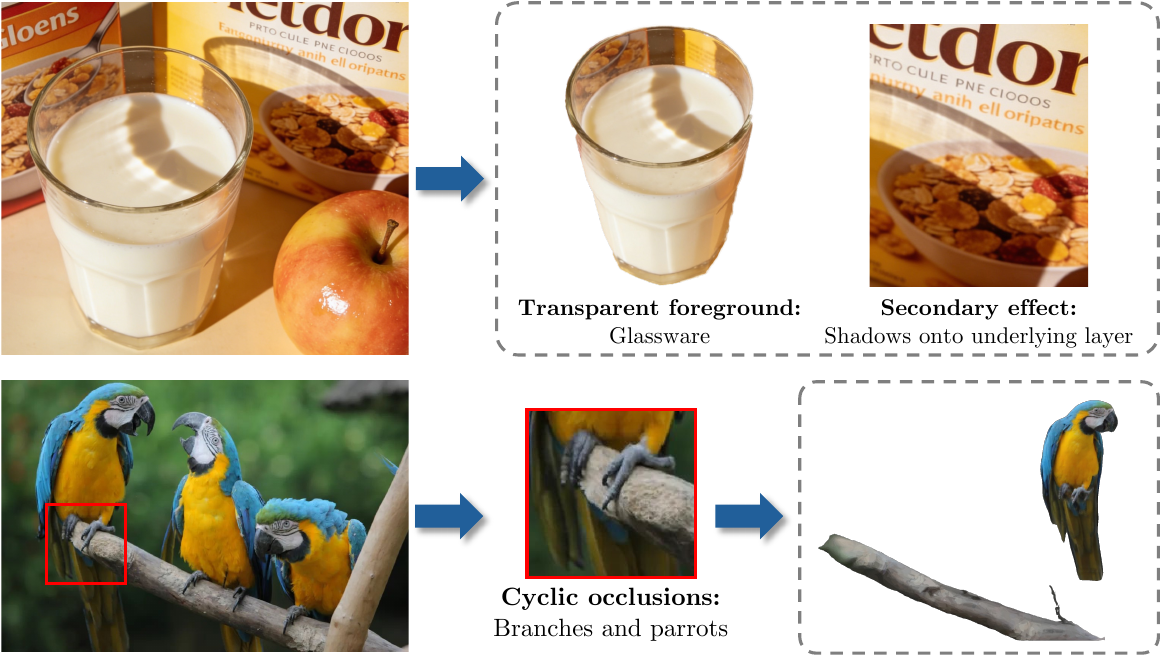}
    \vspace{-1em}
    \caption{\textbf{Failure cases of AmodalSVG.} \textbf{(Top) Transmissive objects and secondary effects.} Peeling transparent entities (\eg, glassware) disrupts natural color blending, resulting in ``baked-in'' foreground colors and ``orphaned'' shadows on the background. \textbf{(Bottom) Cyclic occlusions.} Intertwined objects (\eg, a parrot gripping a branch) create depth cycles that cannot be natively rendered by the strict global z-ordering of the SVG format.}
    \label{fig:supp_fail}
    \vspace{-1.5em}
\end{figure}
While \textbf{AmodalSVG} achieves high-fidelity amodal vectorization, it faces limitations in scenes with complex photometric or topological entanglement (\cref{fig:supp_fail}). First, our layer-wise peeling inherently targets opaque geometry, making it challenged by \textbf{transparent foreground objects} (\eg, glassware) and \textbf{secondary optical effects} (\eg, cast shadows). Separating these entities disrupts transmissive color blending, forcing physical transparency to be unnaturally ``baked'' into the foreground as static colors. Meanwhile, residual effects like shadows are left ``orphaned'' on the background canvas, severely complicating subsequent inpainting and vectorization. Second, \textbf{cyclic occlusions} present a fundamental topological hurdle. In interwoven scenarios—such as a parrot's claws gripping a branch that concurrently occludes its body—our pipeline successfully extracts individual amodal entities but struggles to reconstitute the exact composite. This limitation is inherent to the standard SVG format's \textit{strict global z-ordering}. It cannot represent depth cycles without manually subdividing paths into fragmented sub-components, which directly contradicts our objective of preserving amodal entity integrity.

\end{document}